\documentclass[11pt,letterpaper]{article}
\usepackage[letterpaper]{geometry}
\usepackage{naaclhlt2016}
\usepackage{times}
\usepackage{latexsym}
\usepackage{amsmath}
\usepackage{amssymb}
\usepackage{multirow}
\usepackage{url}
\usepackage{graphicx}
\usepackage{makecell}
\usepackage{subcaption}

\aclfinaltrue
\title{Generating Descriptions from Structured Data Using a Bifocal Attention Mechanism and Gated Orthogonalization}

\author{Preksha Nema$^{\dagger}$\thanks{* The first three authors have contributed equally to this work.} \hspace{0.2cm} Shreyas Shetty$^{\dagger}$\footnotemark[1] \hspace{0.2cm} Parag Jain$^{\bullet}$\footnotemark[1] \\
\bf{Anirban Laha}$^{\bullet\dagger}$ \hspace{0.1cm} Karthik Sankaranarayanan$^{\bullet}$ \hspace{0.1cm}  Mitesh M. Khapra$^{\dagger\ddagger}$\\
  $^\dagger$IIT Madras, India \hspace{0.1cm} 
  $^{\bullet}$IBM Research\\
    $^{\ddagger}$ Robert Bosch Center for Data Science and Artificial Intelligence, IIT Madras \\
  {\tt \{preksha,shshett,miteshk\}@cse.iitm.ac.in} \\ {\tt \{pajain34,anirlaha,kartsank\}@in.ibm.com}
}
\hypersetup{draft}

\begin{document}
\maketitle
\begin{abstract}

In this work, we focus on the task of generating natural language descriptions from a structured table of facts containing \textit{fields} (such as nationality, occupation, \textit{etc}) and \textit{values} (such as Indian, \{actor, director\}, \textit{etc}). One simple choice is to treat the table as a sequence of \textit{fields} and \textit{values} and then use a standard seq2seq model for this task. However, such a model is too generic and does not exploit task-specific characteristics. For example, while generating descriptions from a table, a human would attend to information at two levels: (i) the fields (macro level) and (ii) the values within the field (micro level). Further, a human would continue attending to a field for a few timesteps till all the information from that field has been rendered and then never return back to this field (because there is nothing left to say about it). To capture this behavior we use (i) a fused bifocal attention mechanism which exploits and combines this micro and macro level information and (ii) a gated orthogonalization mechanism which tries to ensure that a field is remembered for a few time steps and then forgotten.  We experiment with a recently released dataset which contains fact tables about people and their corresponding one line biographical descriptions in English. In addition, we also introduce two similar datasets for French and German. Our experiments show that the proposed model gives $21$\% relative improvement over a recently proposed state of the art method and $10$\% relative improvement over basic seq2seq models. The code and the datasets developed as a part of this work are publicly available. \footnote{\url{https://github.com/PrekshaNema25/StructuredData_To_Descriptions}}

\end{abstract}

\section{Introduction}
\begin{figure}[t]
\begin{center}
    \includegraphics[width=0.48\textwidth, height=40mm]{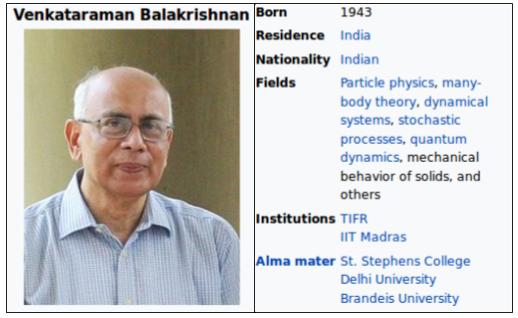}
\end{center}
\caption{Sample Infobox with description : \emph{V. Balakrishnan (born 1943 as Venkataraman Balakrishnan) is an Indian theoretical physicist who has worked in a number of fields of areas, including particle physics, many-body theory, the mechanical behavior of solids, dynamical systems, stochastic processes, and quantum dynamics.}}
\label{fig:vbalki}
\end{figure}

Rendering natural language descriptions from structured data is required in a wide variety of commercial applications such as generating descriptions of products, hotels, furniture, \textit{etc}., from a corresponding table of facts about the entity. Such a table typically contains \{\textit{field, value}\} pairs where the field is a property of the entity (\textit{e.g.}, \textit{color}) and the value is a set of possible assignments to this property (\textit{e.g.}, \textit{color = red}). Another example of this is the recently introduced task of generating one line biography descriptions from a given Wikipedia infobox \cite{lebret2016neural}. The Wikipedia infobox serves as a table of facts about a person and the first sentence from the corresponding article serves as a one line description of the person. Figure \ref{fig:vbalki} illustrates an example input infobox which contains fields such as Born, Residence, Nationality, Fields, Institutions and Alma Mater. Each field further contains some words (\textit{e.g.}, particle physics, many-body theory, \textit{etc}.). The corresponding description is coherent with the information contained in the infobox.

Note that the number of fields in the infobox and the ordering of the fields within the infobox varies from person to person. Given the large size (700K examples) and heterogeneous nature of the dataset which contains biographies of people from different backgrounds (sports, politics, arts, \textit{etc.}), it is hard to come up with simple rule-based templates for generating natural language descriptions from infoboxes, thereby making a case for data-driven models. Based on the recent success of data-driven neural models for various other NLG tasks \cite{bahdanau15,rush15,dialog15,chopra16,diverseattn17}, one simple choice is to treat the infobox as a sequence of \{field, value\} pairs and use a standard seq2seq model for this task. However, such a model is too generic and does not exploit the specific characteristics of this task as explained below.

First, note that while generating such descriptions from structured data, a human keeps track of information at two levels. Specifically, at a macro level, she would first decide which field to mention next and then at a micro level decide which of the values in the field needs to be mentioned next. For example, she first decides that at the current step, the field \textit{occupation} needs attention and then decides which is the next appropriate occupation to attend to from the set of occupations (\textit{actor, director, producer, \textit{etc.}}). To enable this, we use a bifocal attention mechanism which computes an attention over \textit{fields} at a macro level and over \textit{values} at a micro level. We then fuse these attention weights such that the attention weight for a field also influences the attention over the values within it. Finally, we feed a fused context vector to the decoder which contains both field level and word level information. Note that such two-level attention mechanisms \cite{nallapati2016abstractive,hovy16,hred16} have been used in the context of unstructured data (as opposed to structured data in our case), where at a macro level one needs to pay attention to sentences and at a micro level to words in the sentences.

Next, we observe that while rendering the output, once the model pays attention to a field (say, occupation) it needs to stay on this field for a few timesteps (till all the occupations are produced in the output). We refer to this as the \textit{stay on} behavior. Further, we note that once the tokens of a field are referred to, they are usually not referred to later. For example, once all the occupations have been listed in the output we will never visit the occupation field again because there is nothing left to say about it. We refer to this as the \textit{never look back} behavior. To model the \textit{stay on} behaviour, we introduce a forget (or remember) gate which acts as a signal to decide when to forget the current field (or equivalently to decide till when to remember the current field). To model the \textit{never look back} behaviour we introduce a gated orthogonalization mechanism which ensures that once a field is forgotten, subsequent field context vectors fed to the decoder are orthogonal to (or different from) the previous field context vectors.

We experiment with the {\scshape WikiBio} dataset \cite{lebret2016neural} which contains around 700K \{infobox, description\} pairs and has a vocabulary of around 400K words. We show that the proposed model gives a relative improvement of $21$\% and $20$\% as compared to current state of the art models \cite{lebret2016neural,weather16} on this dataset. The proposed model also gives a relative improvement of $10$\% as compared to the basic seq2seq model. Further, we introduce new datasets for French and German on the same lines as the English {\scshape WikiBio} dataset. Even on these two datasets, our model outperforms the state of the art methods mentioned above. 



 

\section{Related work}

Natural Language Generation has always been of interest to the research community and has received a lot of attention in the past. The approaches for NLG range from (i) rule based approaches (\textit{e.g.}, \cite{coral2003,reiter05,green06,galanis07,turner10}) (ii) modular statistical approaches which divide the process into three phases (planning, selection and surface realization) and use data driven approaches for one or more of these phases  \cite{barlapata05,belz08,angeli10,kimmooney10,KonstasL13} (iii) hybrid approaches which rely on a combination of handcrafted rules and corpus statistics  \cite{langkilde98,soricut06,francoiswalker11} and (iv) the more recent neural network based models \cite{bahdanau15}.

Neural models for NLG have been proposed in the context of various tasks such as machine translation \cite{bahdanau15}, document summarization \cite{rush15,chopra16}, paraphrase generation \cite{prakash16},  image captioning \cite{icml2015_xuc15}, video summarization \cite{venu14}, query based document summarization \cite{diverseattn17} and so on. Most of these models are data hungry and are trained on large amounts of data. On the other hand, NLG from structured data has largely been studied in the context of small datasets such as {\scshape WeatherGov} \cite{liang09}, {\scshape RoboCup} \cite{chenmooney08}, {\scshape NFL Recaps} \cite{barlapata05},  {\scshape Prodigy-Meteo} \cite{belz09} and {\scshape TUNA} Challenge \cite{tuna}. Recently \newcite{weather16} proposed RNN/LSTM based neural encoder-decoder models with attention for {\scshape WeatherGov} and {\scshape RoboCup} datasets.


Unlike the datasets mentioned above, 
the biography dataset introduced by \newcite{lebret2016neural} is larger (700K \{table, descriptions\} pairs) and has a much larger vocabulary (400K words as opposed to around 350 or fewer words in the above datasets). Further, unlike the feed-forward neural network based model proposed by \cite{lebret2016neural} we use a sequence to sequence model and introduce components to address the peculiar characteristics of the task. Specifically, we introduce neural components to address the need for attention at two levels and to address the \textit{stay on} and \textit{never look back} behaviour required by the decoder. \newcite{KiddonZC16} have explored the use of checklists to track previously visited ingredients while generating recipes from ingredients. Note that two-level attention mechanisms have also been used in the context of summarization \cite{nallapati2016abstractive}, document classification \cite{hovy16}, dialog systems \cite{hred16}, \textit{etc}. However, these works deal with unstructured data (sentences at the higher level and words at a lower level) as opposed to structured data in our case.

\begin{figure*}
\centering
\includegraphics[width=\textwidth, height=5.5cm]{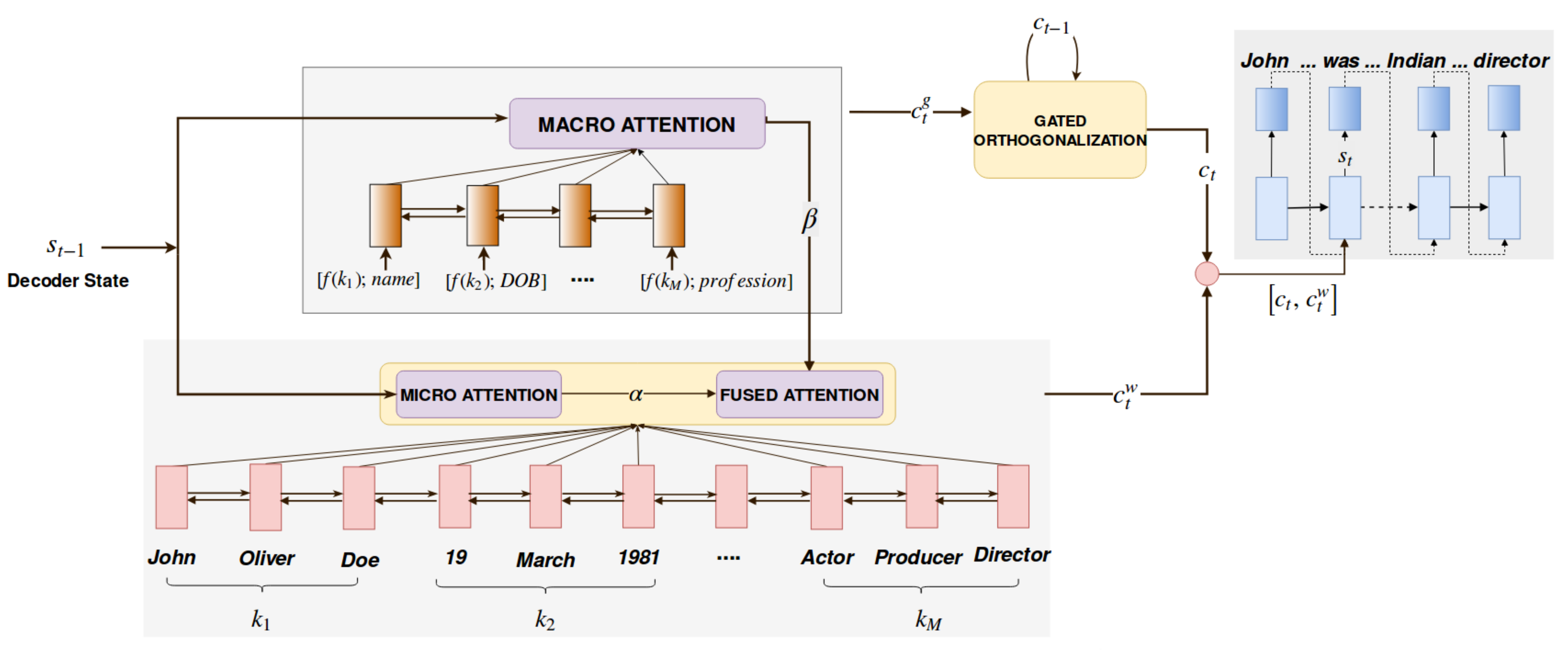}
\caption{Proposed model}
\label{fig:model}
\end{figure*}

\section{Proposed model}

As input we are given an infobox $\mathcal{I} = \{(g_i, k_i) \}_{i=1}^{M}$, which is 
a set of pairs $(g_i, k_i)$ where $g_i$ corresponds to \textit{field} names and $k_i$ is the sequence of corresponding \textit{values} and $M$ is the total number of fields in $\mathcal{I}$. For example, $(g = \textit{occupation}, k=\textit{actor, writer, director})$ could be one such pair in this set. 
Given such an input, the task is to generate a description $y = y_1, y_2, \dots, y_m$ containing $m$ words. A simple solution is to treat the infobox as a sequence of \textit{fields} followed by the \textit{values} corresponding to the field in the order of their appearance in the infobox. For example, the infobox could be flattened to produce the following input sequence (the words in bold are field names which act as delimiters)

\textit{\textbf{[Name]} John Doe \textbf{[Birth\_Date]} 19 March 1981 \textbf{[Nationality]} Indian .....}

The problem can then be cast as a seq2seq generation problem and can be modeled using a standard neural architecture comprising of three components (i) an input encoder (using GRU/LSTM cells), (ii) an attention mechanism to attend to important values in the input sequence at each time step and (iii) a decoder to decode the output one word at a time (again, using GRU/LSTM cells). However, this standard model is too generic and does not exploit the specific characteristics of this task. We propose additional components, \textit{viz.}, (i) a fused bifocal attention mechanism which operates on fields (macro) and values (micro) and (ii) a gated orthogonalization mechanism to model \textit{stay on} and \textit{never look back} behavior.

\subsection{Fused Bifocal Attention Mechanism}
\label{sec:hier}
Intuitively, when a human writes a description from a table she keeps track of information at two levels. At the macro level, it is important to decide which is the appropriate field to attend to next and at a micro level (\textit{i.e.}, within a field) it is important to know which values to attend to next. To capture this behavior, we use a bifocal attention mechanism as described below.

\noindent{\textbf{Macro Attention:}} Consider the $i$-th field $g_i$ which has values $k_i = (w_1, w_2, ..., w_p)$. 
Let $h_i^g$ be the representation of this field in the infobox. This representation can either be (i) the word embedding of the field name or (ii) some function $f$ of the values in the field or (iii) a concatenation of (i) and (ii). The function $f$ could simply be the sum or average of the embeddings of the values in the field. Alternately, this function could be a GRU (or LSTM) which treats these values within a field as a sequence and computes the field representation as the final representation of this sequence (\textit{i.e.}, the representation of the last time-step). We found that bidirectional GRU is a better choice for $f$ and concatenating the embedding of the field name with this GRU representation works best. Further, using a bidirectional GRU cell to take contextual information from neighboring fields also helps (these are the orange colored cells in the top-left block in Figure \ref{fig:model} with macro attention). Given these representations $\{h_i^g\}_{i=1}^{M}$  for all the $M$ fields we compute an attention over the fields (macro level).
\begin{align}
\nonumber b_{t,i}^g &= v_g^T \tanh(U_g s_{t-1} + V_g h^g_i) \\
\nonumber \beta_{t,i} &= \frac{\text{exp}(b_{t,i}^g)}{\sum_{l=1}^{M} \text{exp}(b_{t,l}^g)} \\
c^g_t &= \sum_{i=1}^{M} \beta_{t,i} h_i^g
\label{c}
\end{align}

where $s_{t-1}$ is the state of the decoder at time step $t-1$. $U_g, V_g$ and $v_g$ are parameters, $M$ is the total number of fields in the input, $c^g_t$ is the macro (field level) context vector at the $t$-th time step of the decoder.

\noindent{\textbf{Micro Attention:}}
Let $h_j^{w}$ be the representation of the $j$-th value in a given field. This representation could again either be (i) simply  the embedding of this value (ii) or a contextual representation computed using a function $f$ which also considers the other values in the field. For example, if $(w_1, w_2, ..., w_p)$ are the values in a field then these values can be treated as a sequence and the representation of the $j$-th value can be computed using a bidirectional GRU over this sequence. Once again, we found that using a bi-GRU works better then simply using the embedding of the value. Once we have such a representation computed for all values across all the fields, we compute the attention over these values (micro level) as shown below :

\begin{align}
a_{t,j}^w &= v_w^T \tanh(U_w s_{t-1} + V_w h^w_j) \\
\alpha_{t,j}^w &= \frac{\text{exp}(a_{t,j}^w)}{\sum_{l=1}^{W} \text{exp}(a_{t,l}^w)} 
\end{align}
where $s_{t-1}$ is the state of the decoder at time step $t-1$. $U_w, V_w $ and $v_w$ are parameters, $W$ is the total number of values across all the fields. 

\noindent{\textbf{Fused Attention:}} Intuitively, the attention weights assigned to a field should have an influence on all the values belonging to the particular field. To ensure this, we reweigh the micro level attention weights based on the corresponding macro level attention weights. In other words, we fuse the attention weights at the two levels as:
\begin{align}
\label{eqn_comb_attn}
\alpha_{t,j}^{'} &= \frac{\alpha_{t,j} \beta_{t, F(j)}}{\sum_{l=1}^{W} \alpha_{t,l} \beta_{t, F(l)}} \\
c_t^{w} &= \sum_{j=1}^{W} \alpha^{'}_{t,j} h_j^w  
\end{align}
where $F(j)$ is the field corresponding to the $j$-th value, $c_t^{w}$ is the macro level context vector. 

\subsection{Gated Orthogonalization for Modeling Stay-On and Never Look Back behaviour}  \label{sec:nlb}
We now describe a series of choices made to model \textit{stay-on} and \textit{never look back} behavior. We first begin with the \textit{stay-on} property which essentially implies that if we have paid attention to the field $i$ at timestep $t$ then we are likely to pay attention to the same field for a few more time steps. For example, if we are focusing on the \textit{occupation} field at this timestep then we are likely to focus on it for the next few timesteps till all relevant values in this field have been included in the generated description. In other words, we want to remember the field context vector $c_t^{g}$ for a few timesteps. One way of ensuring this is to use a remember (or forget) gate as given below which remembers the previous context vector when required and forgets it when it is time to move on from that field. 
\begin{align}
f_t &= \sigma(W_{t}^fc_{t-1}^{g} + W_{g}^f c_{t-1} +b_f) \label{eq2}\\
c_t &= (1-f_t)\odot c_{t}^{g} +  f_{t}\odot c_{t-1} \label{eq3}
\end{align}

where ${W_t}^f , {W_g}^f, b_f$  are parameters to be learned. The job of the forget gate is to ensure that $c_{t}$ is similar to $c_{t-1}$ when required (i.e., by learning $f_t \rightarrow 1$ when we want to continue focusing on the same field) and different when it is time to move on (by learning that $f_t \rightarrow 0$). 

Next, the \textit{never look back} property implies that once we have moved away from a field we are unlikely to pay attention to it again. For example, once we have rendered all the occupations in the generated description there is no need to return back to the occupation field. In other words, once we have moved on ($f_t \rightarrow 0$), we want the successive field context vectors $c_{t}^{g}$ to be very different from the previous field vectors $c_{t-1}$. One way of ensuring this is to orthogonalize successive field vectors using 
\begin{align}
\label{eq4}
c_{t}^{g}  = c_{t}^{g} - \gamma_t\odot\frac{<c_{t-1}, c_{t}^{g}>}{<c_{t-1}, c_{t-1}>} c_{t-1}
\end{align}

where $<a,b>$ is the dot product between vectors $a$ and $b$. The above equation essentially subtracts the component of $c_{t}^{g}$ along $c_{t-1}$. $\gamma_t$ is a learned parameter which controls the degree of orthogonalization thereby allowing a soft orthogonalization (\textit{i.e.,} the entire component along $c_{t-1}$ is not subtracted but only a fraction of it). The above equation only ensures that $c_{t}^{g}$ is soft-orthogonal to $c_{t-1}$. Alternately, we could pass the sequence of context vectors, $c_1, c_2, ..., c_t$ generated so far through a GRU cell. The state of this GRU cell at each time step would thus be aware of the history of the field vectors till that timestep. Now instead of orthogonalizing $c_{t}^{g}$ to $c_{t-1}$ we could orthogonalize $c_{t}^{g}$  to the hidden state of this GRU at time-step $t-1$. In practice, we found this to work better as it accounts for all the field vectors in the history instead of only the previous field vector.

In summary, Equation \ref{eq3} provides a mechanism for remembering the current field vector when appropriate (thus capturing \textit{stay-on} behavior) using a remember gate. On the other hand, Equation \ref{eq4} explicitly ensures that the field vector is very different (soft-orthogonal) from the previous field vectors once it is time to move on (thus capturing \textit{never look back} behavior). The value of $c^g_t$ computed in Equation \ref{eq4} is then used in Equation \ref{eq3}. The $c_t$ (macro) thus obtained is then concatenated with $c^w_t$ (micro) and fed to the decoder (see Fig. \ref{fig:model})

\section{Experimental setup}
We now describe our experimental setup: 

\subsection{Datasets}
\label{ds}

We use the {\scshape WikiBio} dataset introduced by \newcite{lebret2016neural}. It consists of $728,321$ biography articles from English Wikipedia. A biography article corresponds to a person (sportsman, politician, historical figure, actor, \textit{etc}.). Each Wikipedia article has an accompanying infobox which serves as the structured input and the task is to generate the first sentence of the article (which typically is a one-line description of the person). We used the same train, valid and test sets which were made publicly available by \newcite{lebret2016neural}. 

We also introduce two new biography datasets, one in French and one in German. These datasets were created and pre-processed using the same procedure as outlined in \newcite{lebret2016neural}. Specifically, we extracted the infoboxes and the first sentence from the corresponding Wikipedia article. As with the English dataset, we split the French and German datasets randomly into train ($80$\%), test ($10$\%) and valid ($10$\%). The French and German datasets extracted by us has been made publicly available.\footnote{\url{https://github.com/PrekshaNema25/StructuredData_To_Descriptions}} The number of examples was 170K and 50K  and the vocabulary size was 297K and 143K for French and German respectively. Although in this work we focus only on generating descriptions in one language, we hope that this dataset will also be useful for developing models which jointly learn to generate descriptions from structured data in multiple languages. 
\subsection{Models compared}

We compare with the following models:

\noindent \textbf{1. \cite{lebret2016neural}:} This is a conditional language model which uses a feed-forward neural network to predict the next word in the description conditioned on \textit{local characteristics} (\textit{i.e.,} words within a field) and \textit{global characteristics}  (\textit{i.e.,} overall structure of the infobox).

\noindent \textbf{2. \cite{weather16}:} This model was proposed in the context of the {\scshape WeatherGov} and {\scshape RoboCup} datasets which have a much smaller vocabulary. 
They use an improved attention model with additional regularizer terms which influence the weights assigned to the fields.


\noindent \textbf{3. Basic Seq2Seq:} This is the vanilla encode-attend-decode model \cite{bahdanau15}. Further,  to deal with the large vocabulary ($\sim$400K words) we use a copying mechanism as a post-processing step. Specifically, we identify the time steps at which the decoder produces unknown words (denoted by the special symbol UNK). For each such time step, we look at the attention weights on the input words and replace the UNK word by that input word which has received maximum attention at this timestep. This process is similar to the one described in \cite{luong2014addressing}. Even \newcite{lebret2016neural} have a copying mechanism tightly integrated with their model.

\subsection{Hyperparameter tuning}
We tuned the hyperparameters of all the models using a validation set. As mentioned earlier, we used a bidirectional GRU cell as the function $f$ for computing the representation of the fields and the values (see Section \ref{sec:hier}). For all the models, we experimented with GRU state sizes of $128$, $256$ and $512$. 
The total number of unique words in the corpus is around 400K (this includes the words in the infobox and the descriptions). Of these, we retained only the top 20K words in our vocabulary (same as \cite{lebret2016neural}). We initialized the embeddings of these words with $300$ dimensional Glove embeddings \cite{pennington2014glove}. 
We used Adam \cite{adam14} with a learning rate of $0.0004$, $\beta_1=0.9$ and $\beta_2=0.999$. We trained the model for a maximum of $20$ epochs and used early stopping with the patience set to $5$ epochs. 

\begin{table}
\resizebox{\columnwidth}{!}{
\begin{tabular}{|l|l|l|l|}
\hline
\textbf{Model}         & \textbf{BLEU-4} & \textbf{NIST-4} & \textbf{ROUGE-4} \\ \hline
\cite{lebret2016neural}        & 34.70   & 7.98   & 25.80    \\ \hline
\cite{weather16}        & 35.10   & 7.27   & 30.90    \\ \hline 
Basic Seq2Seq & 38.20   & 8.47   & 34.28   \\ \hline 
+Fused bifocal attention     & 41.22  & 8.96   & 38.71   \\ \hline 
+Gated orthogonalization  & \textbf{42.03}  & \textbf{9.17}   & \textbf{39.11}  \\ \hline
\end{tabular}
}
\caption{Comparison of different models on the English {\scshape WikiBio} dataset}
\label{tab:wbio-eng}
\end{table}


\begin{table*}[h]
    \footnotesize
    \begin{tabular}{|p{15.4cm}|}
        \hline
        \textbf{Reference:} Samuel Smiles (23 December 1812 –- 16 April 1904), was a Scottish author and government reformer who campaigned on a Chartist platform.\\
        \textbf{Basic Seq2Seq:} samuel smiles (23 december 1812 -- 16 april 1904) was an english books and author. \\
        \textbf{+Bifocal attention:} samuel smiles (23 december 1812 - 16 april 1904) was a british books and books. \\
        \textbf{+Gated Orthogonalization:} samuel smiles (23 december 1812 - 16 april 1904) was a british biographies and author.\\
        \hline
        \textbf{Reference:} Thomas Tenison (29 September 1636 –- 14 December 1715) was an English church leader, Archbishop of Canterbury from 1694 until his death. \\
        \textbf{Basic Seq2Seq:} thomas tenison (14 december 1715 - 29 september 1636) was an english roman catholic archbishop. \\
        \textbf{+Bifocal attention:} thomas tenison (29 september 1636 - 14 december 1715) was an english clergyman of the roman catholic church.\\
        \textbf{+Gated Orthogonalization:} thomas tenison (29 september 1636 - 14 december 1715) was archbishop of canterbury from 1695 to 1715. \\
        \hline
        \textbf{Reference:} Guy F. Cordon (April 24, 1890 –- June 8, 1969) was a U.S. politician and lawyer from the state of Oregon. \\
        \textbf{Basic Seq2Seq:} charles l. mcnary (april 24 , 1890 8 , 1969) was a united states senator from oregon. \\
        \textbf{+Bifocal attention:}guy cordon (april 24 , 1890 -- june 8 , 1969) was an american attorney and politician.\\
        \textbf{+Gated Orthogonalization:} guy cordon (april 24 , 1890 -- june 8 , 1969) was an american attorney and politician from the state of oregon. \\
        \hline
        \textbf{Reference:} Dr. Harrison B. Wilson Jr. (born April 21, 1925) is an American educator and college basketball coach who served as the second president of Norfolk State University from 1975-1997. \\
        \textbf{Basic Seq2Seq:} lyman beecher brooks (born april 21 , 1925) is an american educator and educator. \\
        \textbf{+Bifocal attention:} harrison b. wilson , jr. (born april 21 , 1925) is an american educator and academic administrator.  \\
        \textbf{+Gated Orthogonalization:} harrison b. wilson , jr. (born april 21 , 1925) is an american educator , academic administrator , and former president of norfolk state university. \\
        \hline
    \end{tabular}
    \caption{Examples of generated descriptions from different models. For the last two examples, $name$ generated by Basic Seq2Seq model is incorrect because it attended to \textit{preceded by} field.}
    \label{tab:samp_desc}
\end{table*}

\begin{table}[!h]
\centering
\begin{tabular}{|c|c|c|c|}
\hline
\textbf{Metric} & A \textless B & A == B & A \textgreater B \\ \hline
\textbf{Adequacy}   & 186                                                                            & 208                                                                      & 106                                                                              \\
\textbf{Fluency}    & 244                                                                            & 108                                                                      & 148                                                                              \\
\textbf{Preference} & 207                                                                            & 207                                                                      & 86                                                                               \\ \hline
\end{tabular}
\caption{Qualitative Comparison of Model A (Seq2Seq) and Model B (our model)}
\label{tab:human_eval}
\end{table}

\section{Results and Discussions}
We now discuss the results of our experiments.


\subsection{Comparison of different models}
Following \newcite{lebret2016neural}, we used BLEU-4, NIST-4 and ROUGE-4 as the evaluation metrics. We first make a few observations based on the results on the English dataset (Table \ref{tab:wbio-eng}). 
The basic seq2seq model, as well as the model proposed by \newcite{weather16}, perform better than the model proposed by \newcite{lebret2016neural}. Our final model with bifocal attention and gated orthogonalization gives the best performance and does $10$\% (relative) better than the closest baseline (basic seq2seq) and $21$\% (relative) better than the current state of the art method \cite{lebret2016neural}. In Table \ref{tab:samp_desc}, we show some qualitative examples of the output generated by different models.  

\subsection{Human Evaluations}
To make a qualitative assessment of the generated sentences, we conducted a human study on a sample of 500 Infoboxes which were sampled from English dataset. The annotators for this task were undergraduate and graduate students. For each of these infoboxes, we generated summaries using the basic seq2seq model and our final model with bifocal attention and gated orthogonalization. For each description and for each model, we asked three annotators to rank the output of the systems based on i) adequacy (\textit{i.e.} does it capture relevant information from the infobox), (ii) fluency (\textit{i.e.} grammar) and (iii) relative preference (\textit{i.e.}, which of the two outputs would be preferred). Overall the average fluency/adequacy (on a scale of $5$) for basic seq2seq model was $4.04/3.6$ and $4.19/3.9$  for our model respectively.

The results from Table \ref{tab:human_eval} suggest that in general gated orthogonalization model performs better than the basic seq2seq model. Additionally, annotators were asked to verify if the generated summaries look natural (\textit{i.e}, as if they were generated by humans). In $423$ out of $500$ cases, the annotators said ``Yes'' suggesting that gated orthogonalization model indeed produces good descriptions.

\begin{table}
\resizebox{1\columnwidth}{!}{
\begin{tabular}{|l|l|l|l|}
\hline
\textbf{Model}         & \textbf{BLEU-4} & \textbf{NIST-4} & \textbf{ROUGE-4} \\ \hline
\cite{weather16}  & 10.40   & 2.51   & 7.81    \\ \hline
Basic Seq2Seq & 14.50   & 3.02   & 12.22   \\ \hline
+Fused bifocal attention     & 13.80   & 2.86   & 12.37   \\ \hline
+Gated orthogonalization        & \textbf{15.52}  & \textbf{3.30}   & \textbf{12.80}    \\ \hline
\end{tabular}
}
\caption{Comparison of different models on the French {\scshape WikiBio} dataset}
\label{tab:frwbio}

\resizebox{\columnwidth}{!}{
\begin{tabular}{|l|l|l|l|}
\hline
\textbf{Model}         & \textbf{BLEU-4} & \textbf{NIST-4} & \textbf{ROUGE-4} \\ \hline
\cite{weather16}  & 9.30   & 2.23   & 5.85    \\ \hline
Basic Seq2Seq & 17.05   & 3.09   & 12.16   \\ \hline
+Fused bifocal attention & 20.38   & 3.43   & 14.89   \\ \hline
+Gated orthogonalization        & \textbf{23.33}  & \textbf{4.24}   & \textbf{16.40} \\ \hline
\end{tabular}
}
\caption{Comparison of different models on the German {\scshape WikiBio} dataset}
\label{tab:dewbio}
\end{table}

\begin{figure*}
    \centering
    \begin{subfigure}{0.48\textwidth}
        \centering
\includegraphics[width=\textwidth, height=70mm]{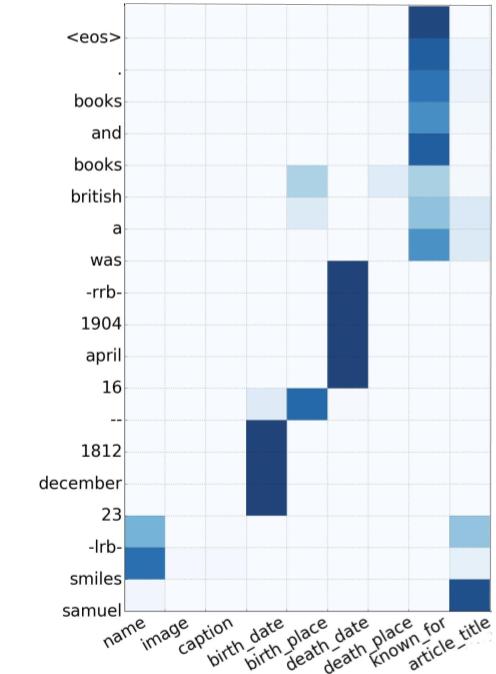}
\caption{Fused Bifocal Attention}
\label{en:q}

    \end{subfigure} %
    ~ 
    \begin{subfigure}{0.48\textwidth}
        \centering
\includegraphics[width=\textwidth, height=70mm]{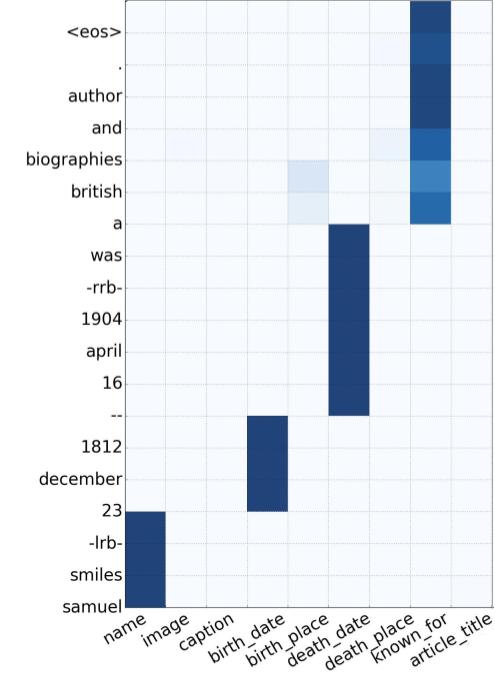}
\caption{Fused Bifocal Attention + Gated Orthogonalization}
\label{en:d}
    \end{subfigure}
    \caption{Comparison of the attention weights and descriptions produced for Infobox in Figure \ref{fig: samuel_smiles}}
    \label{fig:en_hmap}
\end{figure*}

\subsection{Performance on different languages} The results on the French and German datasets are summarized in Tables \ref{tab:frwbio} and \ref{tab:dewbio} respectively. Note that the code of \cite{lebret2016neural} is not publicly available,  hence we could not report numbers for French and German using their model. We observe that our final model gives the best performance - though the bifocal attention model performs poorly as compared to the basic seq2seq model on French. However, the overall performance for French and German are much smaller than those for English. There could be multiple reasons for this. First, the amount of training data in these two languages is smaller than that in English. Specifically, the amount of training data available in French (German) is only $24.2$ ($7.5$)\% of that available for English. Second, on average the descriptions in French and German are longer than that in English (EN: $26.0$ words, FR: $36.5$ words and DE: $32.3$ words). Finally, a manual inspection across the three languages suggests that the English descriptions have a more consistent structure than the French descriptions. For example, most English descriptions start with \textit{name} followed by \textit{date of birth} but this is not the case in French. However, this is only a qualitative observation and it is hard to quantify this characteristic of the French and German datasets.

\begin{figure}
    \centering
    \includegraphics[scale=0.4]{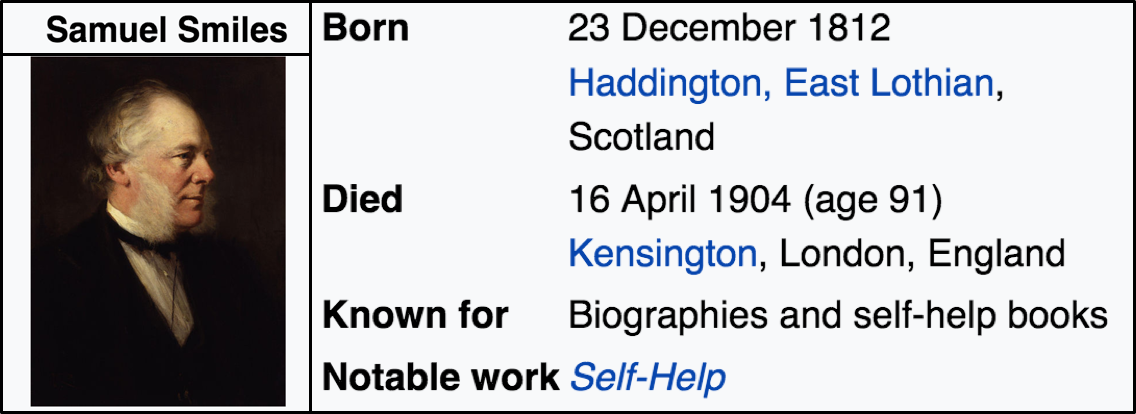}
    \caption{Wikipedia Infobox for Samuel Smiles }
    \label{fig: samuel_smiles}
\end{figure}

\subsection{Visualizing Attention Weights} If the proposed model indeed works well then we should see attention weights that are consistent with the \textit{stay on} and \textit{never look back} behavior. 
To verify this, we plotted the attention weights in cases where the model with gated orthogonalization does better than the model with only bifocal attention. Figure \ref{fig:en_hmap} shows the attention weights corresponding to infobox in Figure \ref{fig: samuel_smiles}. Notice that the model without gated orthogonalization has attention on both name field and article title while rendering the name. The model with gated orthogonalization, on the other hand, stays on the name field for as long as it is required but then moves and never returns to it (as expected).

\begin{figure}
    \centering
    \includegraphics[scale=0.26]{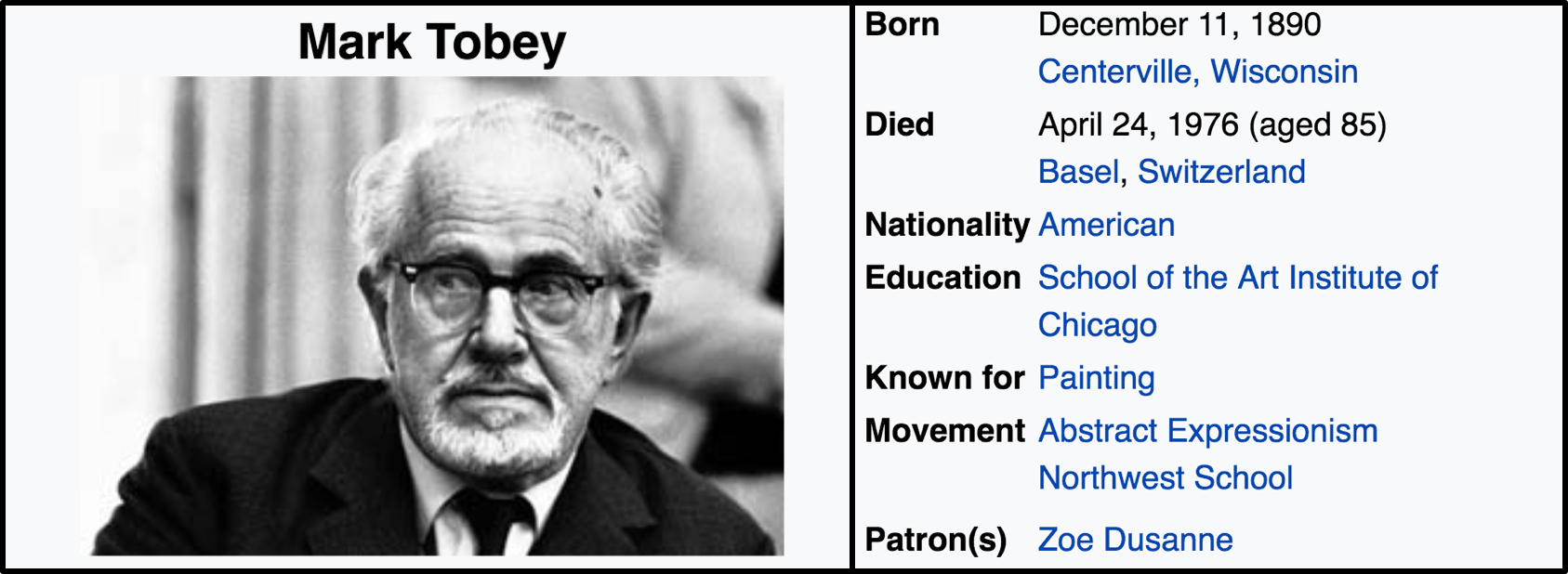}
    \caption{Wikipedia Infobox for Mark Tobey }
    \label{fig: mark_tobey}
\end{figure}

Due to lack of space, we do not show similar plots for French and German but we would like to mention that, in general, the differences between the attention weights learned by the model with and without gated orthogonalization were more pronounced for the French/German dataset than the English dataset. This is in agreement with the results reported in Table \ref{tab:frwbio} and \ref{tab:dewbio} where the improvements given by gated orthogonalization are more for French/German than for English. 



\begin{figure*}   
    \begin{subfigure}{0.5\textwidth}
\includegraphics[width=\textwidth, height=70mm]{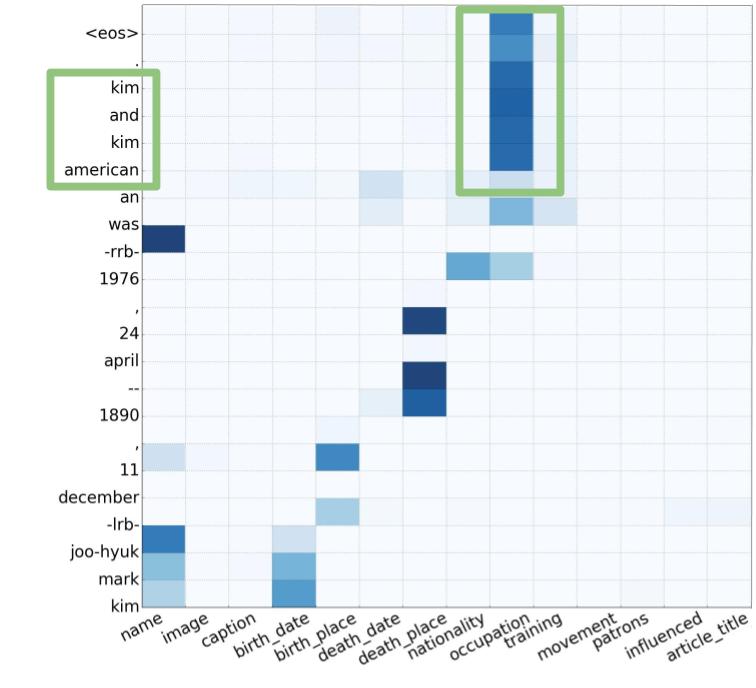}
\caption{Without fine tuning.}
\label{dom:q}
    \end{subfigure}%
    ~ 
    \begin{subfigure}{0.5\textwidth}
\includegraphics[width=\textwidth, height=70mm]{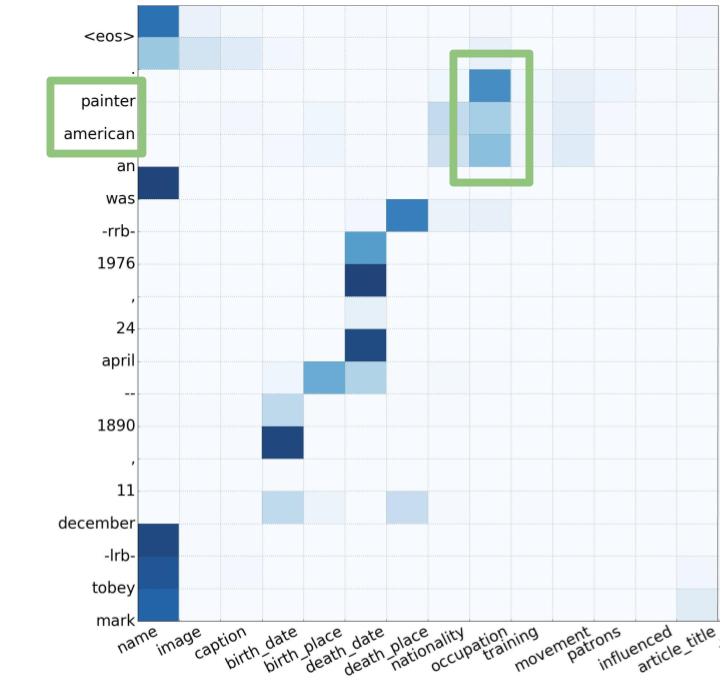}
\caption{With fine tuning with 5K in-domain data.}
\label{dom:d}
    \end{subfigure}
    \caption{Comparison of the attention weights and descriptions (see highlighted boxes) produced by an out-of-domain model with and without fine tuning for the Infobox in Figure \ref{fig: mark_tobey} }
    \label{dom}
\end{figure*}

\subsection{Out of domain results}

\begin{table}
\centering
\footnotesize{
\begin{tabular}{|l|c|c|}
\hline
\multirow{2}{*}{\textbf{Training data}} & \multicolumn{2}{l|}{\textbf{Target (test) data}}     \\ \cline{2-3} 
                                                 & \multicolumn{1}{l|}{\textbf{Arts}} & \textbf{Sports} \\ \hline
Entire dataset                                   & 33.6                               & 52.4            \\ \hline
Without target domain data                       & 24.5                               & 29.3            \\ \hline
+5k target domain data                           & 31.2                               & 41.8            \\ \hline
+10k target domain data                          & 32.2                               & 43.3            \\ \hline
\end{tabular}}
\caption{Out of domain results(BLEU-4)}
\label{tab:dom_adap}
\end{table}

What if the model sees a different $type$ of person at test time? For example, what if the training data does not contain any sportspersons but at test time we encounter the infobox of a sportsperson. This is the same as seeing out-of-domain data at test time. Such a situation is quite expected in the products domain where new products with new features (fields) get frequently added to the catalog.  We were interested in three questions here. First, we wanted to see if testing the model on out-of-domain data indeed leads to a drop in the performance. For this, we compared the performance of our best model in two scenarios (i) trained on data from all domains (including the target domain) and tested on the target domain (sports, arts) and (ii) trained on data from all domains except the target domain and tested on the target domain. Comparing rows 1 and 2 of Table \ref{tab:dom_adap} we observed a significant drop in the performance. Note that the numbers for sports domain in row 1 are much better than the Arts domain because roughly 40\% of the {\scshape WikiBio} training data contains sportspersons.

Next, we wanted to see if we can use a small amount of data from the target domain to fine tune a model trained on the out of domain data. We observe that even with very small amounts of target domain data the performance starts improving significantly (see rows 3 and 4 of Table \ref{tab:dom_adap}). Note that if we train a model from scratch with only limited data from the target domain instead of fine-tuning a model trained on a different source domain then the performance is very poor. In particular, training a model from scratch with 
10K training instances we get a BLEU score of $16.2$ and $28.4$ for arts and sports respectively. Finally, even though the actual words used for describing a sportsperson (footballer, cricketer, \textit{etc.}) would be very different from the words used to describe an artist (actor, musician, \textit{etc}.) they might share many fields (for example, date of birth, occupation, \textit{etc}.). As seen in Figure \ref{dom} (attention weights corresponding to the infobox in Figure \ref{fig: mark_tobey}), the model predicts the attention weights correctly for common fields (such as occupation) but it is unable to use the right vocabulary to describe the occupation (since it has not seen such words frequently in the training data). However, once we fine tune the model with limited data from the target domain we see that it picks up the new vocabulary and produces a correct description of the occupation.

\section{Conclusion}
We present a model for generating natural language descriptions from structured data. To address specific characteristics of the problem we propose neural components for fused bifocal attention and gated orthogonalization to address \textit{stay on} and \textit{never look back} behavior while decoding. 
Our final model outperforms an existing state of the art model on a large scale {\scshape WikiBio} dataset by $21$\%.  We also introduce datasets for French and German and demonstrate that our model gives state of the art results on these datasets. Finally, we perform experiments with an out-of-domain model and show that if such a model is fine-tuned with small amounts of in domain data then it can give an improved performance on the target domain.

Given the multilingual nature of the new datasets, as future work, we would like to build models which can jointly learn to generate natural language descriptions from structured data in multiple languages. One idea is to replace the concepts in the input infobox by Wikidata concept ids which are language agnostic. A large amount of input vocabulary could thus be shared across languages thereby facilitating joint learning.

\section{Acknowledgements}
We thank Google for supporting Preksha Nema through their Google India Ph.D. Fellowship program. We also thank Microsoft Research India for supporting Shreyas Shetty through their generous travel grant for attending the conference.

\bibliography{tacl}
\bibliographystyle{naaclhlt2016}

\end{document}